\documentclass[sigconf]{acmart}
\AtBeginDocument{%
  \providecommand\BibTeX{{%
    \normalfont B\kern-0.5em{\scshape i\kern-0.25em b}\kern-0.8em\TeX}}}

\setcopyright{acmcopyright}
\copyrightyear{2018}
\acmYear{2018}
\acmDOI{10.1145/1122445.1122456}

\acmConference[San Diego '20]{The 26th ACM SIGKDD Conference on Knowledge Discovery and Data Mining (KDD'20)}{Aug. 22--27, 2020}{San Diego, CA}
\acmBooktitle{The 26th ACM SIGKDD Conference on Knowledge Discovery and Data Mining (KDD'20),
  Aug 22--27, 2020, San Diego, CA}
\acmPrice{15.00}
\acmISBN{978-1-4503-XXXX-X/18/06}
\newcommand{\nobracket}{}
\newcommand{\tmmathbf}[1]{\ensuremath{\boldsymbol{#1}}}
\newcommand{\tmop}[1]{\ensuremath{\operatorname{#1}}}
\usepackage{microtype}
\usepackage{graphicx}
\usepackage{subfigure}
\usepackage{booktabs} 
\usepackage{amsmath}
\usepackage{amssymb}
\usepackage{amsfonts}
\usepackage{algorithm}
\usepackage{algorithmic}
\usepackage{float}

\copyrightyear{2020} 
\acmYear{2020} 
\setcopyright{acmcopyright}\acmConference[KDD '20]{Proceedings of the 26th ACM SIGKDD Conference on Knowledge Discovery and Data Mining}{August 23--27, 2020}{Virtual Event, CA, USA}
\acmBooktitle{Proceedings of the 26th ACM SIGKDD Conference on Knowledge Discovery and Data Mining (KDD '20), August 23--27, 2020, Virtual Event, CA, USA}
\acmPrice{15.00}
\acmDOI{10.1145/3394486.3403117}
\acmISBN{978-1-4503-7998-4/20/08}
\begin{document}

\title[ASGN for Molecular Property Prediction]{ASGN: An Active Semi-supervised  Graph Neural Network for Molecular Property Prediction}

\author{Zhongkai Hao$^1$, Chengqiang Lu$^1$,Zhenya Huang$^1$ , Hao Wang$^1$,Zheyuan Hu$^1$, Qi Liu$^{1,*}$,  Enhong Chen$^1$, Cheekong Lee$^2$}
\email{{hzk171805, lunar, huangzhy, wanghao3, ustc_hzy  }@mail.ustc.edu.cn}
\email{{qiliuql, cheneh}@ustc.edu.cn, cheekonglee@tencent.com}
\affiliation{
\institution{ 1: Anhui Province Key Lab of Big Data Analysis and Application, School of Computer Science and Technology, University of Science and Technology of China, 2: Tencent America,
}
}
\renewcommand{\shortauthors}{Zhongkai Hao, et al.}

\begin{abstract}
	Molecular property prediction (e.g., energy) is an essential problem in chemistry and biology. Unfortunately, many supervised learning methods usually suffer from the problem of scarce labeled molecules in the chemical space, where such property labels are generally obtained by Density Functional Theory (DFT) calculation which is extremely computational costly. An effective solution is to incorporate the unlabeled molecules in a semi-supervised fashion. However, learning semi-supervised representation for large amounts of molecules is challenging, including the joint representation issue of both molecular essence and structure, the conflict between representation and property leaning. Here we propose a novel framework called Active Semi-supervised Graph Neural Network (ASGN) by incorporating both labeled and unlabeled molecules. Specifically, ASGN adopts a teacher-student framework. In the teacher model, we propose a novel semi-supervised learning method to learn general representation that jointly exploits information from molecular structure and molecular distribution. Then in the student model, we target at property prediction task to deal with the learning loss conflict.  At last, we proposed a novel active learning strategy in terms of molecular diversities to select informative data during the whole framework learning. We conduct extensive experiments on several public datasets. Experimental results show the remarkable performance of our ASGN framework.

\end{abstract}


\keywords{Active Learning; Molecular Property Prediction; Graph Neural Network; 
Semi-Supervised Learning
\let\thefootnote\relax\footnotetext{$^*$Corresponding Author.}}

\maketitle

{\fontsize{8pt}{8pt} \selectfont
	\textbf{ACM Reference Format:}\\
	Zhongkai Hao, Chengqiang Lu, Zhenya Huang, Hao Wang, Zheyuan Hu, Qi Liu, Enhong Chen, Cheekong Lee. 2020. ASGN: An Active Semi-supervised Graph Neural Network for Molecular Property Prediction. In \textit{Proceedings of the 26th ACM SIGKDD Conference on Knowledge Discovery and Data Mining (KDD '20), August 23-27, 2020, Virtual Event, CA, USA.} ACM, New York, NY, USA, 9 pages.
	https://doi.org/10.1145/3394486.3403117
}

\vspace{0.4cm}
\section{Introduction}

\begin{figure}
	\includegraphics[width=6.5cm,height=6.5cm,trim=100 20 200 30,page=7]{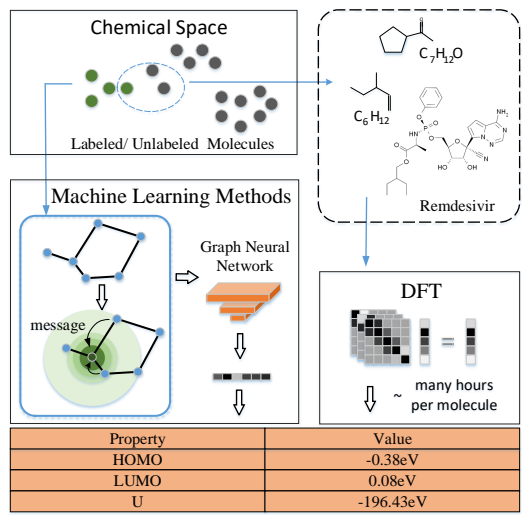}
	\caption{Methods for molecular property prediction. Left: Machine learning methods using message passing graph neural networks. Right: DFT calculation.}
	\vspace{-0.8cm}
	\label{fig1}
	
\end{figure}

Predicting the property of molecules, such as the energy, is a fundamental issue in many related domains including chemistry, biology and material science, which has led to many significant relevant research and applications. For example, the process of drug discovery \cite{ekins2019exploiting} can be accelerated if we can accurately predict the properties of molecules in time to help develop specific medicines for the epidemic, such as H1N1 flu, SARS, Covid19.

In chemistry, density functional theory (DFT) is commonly used computational methods for molecular property prediction, which has been studied dating back to the 1970s~\cite{becke2014perspective}. It  offers accurate and explainable solutions for molecular following complete theory~\cite{kohn1965self}.
However, in practice, it suffers from a critical problem of expensive computation cost as it needs to solve many linear equations iteratively for the solutions. For example, experimental results find that it takes an hour to calculate the properties of a molecule with only 20 atoms \cite{gilmer2017neural}. Obviously, such low efficiency of DFT has limited its applications when screening from a large set of molecules.

Recently, researchers have attempted to use machine learning methods that are cost-effective for molecular property prediction~\cite{hansen2015machine}. Along this line, the most representative methods are graph neural networks (GNN), including MPNN \cite{gilmer2017neural}, SchNet \cite{schutt2017schnet} and MGCN \cite{lu2019molecular}, which have shown superior performance. Generally, they treat a molecule as a graph where the nodes denote atoms and the edges represent the interaction between atoms. They design several neural layers to project each node into latent space with a low-dimensional learnable embedding vector and pass its interaction message through the edges iteratively. At last, the node messages can be aggregated to represent the molecule for property prediction.

Though GNNs have achieved great success, they are usually data-hungry, which requires a big amount of labeled data (i.e., molecules whose properties are known) for training \cite{gilmer2017neural}. However, the labeled molecules usually take an extreme small portion in the whole chemical space since they can only be provided by expensive experiments or DFT calculation, which restricts GNN based development. To gain further promotion, as shown in top left part of the Figure~\ref{fig1}, there are still many valid molecules in the chemical space, though the properties remaining unknown, that have some benefits in terms of their structures. If we can effectively leverage these unlabeled molecules, it could be potentially helpful to improve the performance. Therefore, in this paper, we aim to explore semi-supervised learning (SSL) by fully taking advantage of both labeled molecules and unlabeled ones for property prediction.

However, it is highly challenging due to the following domain-specific characteristics.
First, learning molecular graph representation is non-trivial because it involves both the node and the graph level information. Different from traditional applications like social networks since we usually meet a large number of graphs in chemical space rather than a single graph with large number of nodes. Though some existing semi-supervised learning methods, such as Ladder Networks \cite{rasmus2015semi}, have shown their performance in various domains, such as image and text, they cannot be directly used for molecular graph learning.
Second, it is difficult to handle the imbalance between labeled and unlabeled molecules in chemical space since the number of labeled ones generally take extreme small portion. Directly applying previous SSL methods leads to loss conflict caused by large number of unlabeled molecules for their structural representation but ignores our main goal of property prediction. 
Third, the performance might be still unsatisfactory due to limited labels,
we need to find new molecules for labeling  to improve the
model. To increase the efficiency of labeling, we need a mechanism
to find most informative molecules for labeling.

To address these challenges, we design 
a novel framework called Active Semi-supervised Graph Neural Network (ASGN) for molecular property prediction by taking advantage of both labeled and unlabeled molecules. Generally, ASGN uses a novel teacher-student framework consisting of two models that work alternatively. Specifically, in the teacher model, we propose a novel semi-supervised learning method to learn a general representation that jointly explores molecular features both at a global scale and local scale. The local one represents the essences of molecules, i.e., atoms and bonds while the global one learns the whole molecular graph encoding with respect to the chemical space.
Then, to deal with the loss conflict between the unsupervised structure representation and property prediction, we introduce the student model by fine-tuning on property prediction task only on the small labeled molecules. By doing so, the student model can focus on the prediction to achieve lower error than the teacher model and converge much faster. Additionally, it can alleviate over-fitting than training from scratch only on the labeled dataset. Moreover, to improve labeling efficiency, we propose a novel strategy based on active learning to select new informative molecules. That is, ASGN uses the embeddings by the teacher model to select a diversified subset of molecules in the chemical space and add them to the labeled dataset for finetuning two models repeatedly until the label budget or desired accuracy is reached. We conduct extensive experiments on real-world datasets, where the experimental results demonstrate the effectiveness of our proposed ASGN. To the best of our knowledge, this is the first attempt to incorporate both unlabeled and labeled molecules for property prediction actively in a semi-supervised manner.

\vspace{-0.3cm}
\section{Related Work}
In this section, we summarize the related work with the following three categories.

\textbf{Molecular Property Prediction}.
Predicting the properties of molecules is a fundamental task with applications in many areas such as chemistry and biology \cite{becke2007quantum, oglic2017active}. According to quantum physics, the states of a molecule are characterized by Schrödinger equation \cite{thouless2014quantum}. 
The first class like Density Functional Theory (DFT) \cite{becke2014perspective} 
are simulation based methods directly derived or approximated by the Schrödinger equation.  However, DFT methods are time-consuming because it solves some big linear equations and the complexity of DFT is $O(N^4)$ where $N$ is the number of atoms. 

Another class of molecular properties prediction methods are data-driven \cite{hansen2015machine, ying2018graph, gilmer2017neural, do2019graph}. Researchers attempted to use traditional machine learning methods with empirical descriptors or handcraft features to represent a molecule and use them for linear or logistic regression \cite{hansen2015machine, ying2018graph}. However, these methods cannot achieve desirable accuracy due to the limited effectiveness of handcrafted features and model capacity \cite{gilmer2017neural}.

Inspired by the remarkable development of graph neural networks in various domains  \cite{gilmer2017neural} \cite{wang2019mcne}\cite{DBLP:conf/iclr/PeiWCLY20}\cite{wang2018united}, researchers have noticed the potentials of them for molecular property prediction. Generally, by treating the molecule as a graph, several graph neural networks have been applied  \cite{hamilton2017inductive, ma2019graph, wang2019mcne} as an architecture that can directly deal with noneuclidean data like graphs. Variants of graph neural networks like MPNN \cite{gilmer2017neural}, Schnet\cite{schutt2017schnet}, can be applied for molecular properties prediction where they use nodes to represent atoms, and the edges are weighted by the distances between atoms.
Then the node embeddings are propagated and updated using the embeddings of their neighborhood, named message passing. The graph embedding can be pooled from nodes for property prediction.

\textbf{Semi-supervised Representation learning}.
Semi-supervised learning is a popular framework to improve model performance by incorporating unlabeled data into training \cite{zhu2005semi}. The main idea is to use the unlabeled data to learn a general and robust representation to improve the performance of the model. On the one hand,  methods like ladder network \cite{rasmus2015semi} borrow the idea of jointly learning representation for unlabeled data (via generation) and labeled data \cite{kingma2013auto}. 
On the other hand, a popular fashion is developed recently which uses self-supervised methods that force the networks to be consistent under the handcrafted transformations like image in-painting \cite{pathak2016context}, rotation\cite{gidaris2018unsupervised}, contrastive loss \cite{He}. Usually, these methods use a pseudo-labeling mechanism to assign each unlabeled data with a pseudo label and force the neural network to predict these pesudo labels. Then the pre-trained models can be used for downstream tasks like classification or regression.
For example, \citeauthor{gidaris2018unsupervised} uses the rotation degree of an image as a kind of pesudo label. These pesudo labels are often obtained from transformations of data without changing their semantic feature. Deep Clustering \cite{caron2018deep} shows that the convolutional neural network itself can be viewed as a strong prior to processing image data. Accordingly, they design a self-supervised method based on learning the clustering results of the features by the neural networks.

\textbf{Active Learning}.
Active learning is a popular framework to alleviate data deficiency and it has been applied in many tasks \cite{gal2017deep, yang2014active, DBLP:conf/aaai/WuLZPLC20, DBLP:journals/tois/HuangLCWXCMH20}.
Active learning framework starts with a small set of labeled data and a large set of unlabeled data. In every iteration, it develops a model to select a batch of unlabeled data to be labeled for supplementing the limited labeled data so that it achieves better performance.
Generally, the representative methods consider the strategy selection from two perspectives, i.e., uncertainty, and diversity \cite{gal2017deep} \cite{Sener2017}. Specifically, the uncertainty based methods  define the model uncertainty for a new unlabeled data leveraged by some statistics properties (e.g., variance) and then select the data with the highest value \cite{gal2017deep} \cite{ting2018optimal}. Comparatively, the diversity based methods aim to choose a small subset that is the most representative for the whole dataset \cite{Sener2017}.

As is pointed out in \cite{ash2019deep}, the data selected by the uncertainty strategy are almost identical in batch mode settings, so it might be not suitable for large datasets like our scenarios. In this paper, we propose a novel diversity based active learning strategy for informative molecule selection where the semi-supervised embeddings are used for calculating the distance between molecules.

\section{Definitions and Notations}

In this section, we will give formal definitions of terminologies and problems in this paper for clarity. Following the previous works \cite{gilmer2017neural} \cite{schutt2017schnet}, we treat each molecule in chemical space as a graph, hence we define a molecular graph as follows:
\vspace{-0.12cm}
\begin{definition}
	\textit{\textbf{Molecular Graph}}:
	A molecule is denoted as a weighted graph $\mathcal{G}=(\mathcal{V}, \mathcal{E})$, where the vertex set $\mathcal{V}=\{v_i:1\leq i \leq |\mathcal{G}|\}$, we use $\boldsymbol{x_i}$ to represent the feature vector of the node (atom) $v_i$ indicating its type such as Carbon, Nitrogen. $|\mathcal{G}|$ is the total number of atoms.  $\mathcal{E}=\{e_{i j}=|\boldsymbol{r}_i-\boldsymbol{r}_j|: 1\leq i, j\leq |\mathcal{G}|\}$ is the set of edges connecting two atoms (nodes) $v_i$ and $v_j$. Specifically, in a certain molecule, the coordinates of each atom can be represented as  $\boldsymbol{r}_i=(r^{(1)}_i, r^{(2)}_i, r^{(3)}_i)$.
	Therefore, we further denote the edge between two atom nodes $e_{ij}$ as weighted by their coordinate distance $|r_i - r_j|$.
	
\end{definition}

Then we give the formal definition of chemical space.
\vspace{-0.12cm}
\begin{definition}
	\textit{\textbf{Chemical Space}}:
	Generally, the whole chemical space consists of a set of molecules, which can be denoted as: $M = \{\mathcal{G}_i: 1\leq i \leq N\}$. In practice, only a subset of molecules in the space have been examined to obtain their several properties (e.g., energy) by typical DFT calculation. Therefore, we divide the chemical space $\mathcal{M}$ into two subset $\mathcal{D}_l$, $\mathcal{D}_u$. Specifically, $\mathcal{D}_l=\{(\mathcal{G}_1, \boldsymbol{y}_1),\cdots, (\mathcal{G}_{N_l},\boldsymbol{y}_{N_l})\}$ represents the subset of molecules whose properties have been examined, where $\boldsymbol{y}_i \in \mathbb{R}^m$ denotes the property vector with real value of molecule $\mathcal{G}_i$. Comparatively, $\mathcal{D}_u = \{\mathcal{G}_1, \mathcal{G}_2, \cdots, \mathcal{G}_{N_u}\}$ represents the subset of molecules whose properties remain unknown. Without loss of generality, we call the subset $\mathcal{D}_l$ and $\mathcal{D}_u$ as "labeled set" and "unlabeled set", respectively.
	
\end{definition}

With the above definition, our problem can be formalized as that we want to find a model $f(\mathcal{G}) \to \boldsymbol{y}$ using limited labels $|\mathcal{D}_u|$, for precisely predicting the properties of molecules.

\begin{figure*}
	\centering
	\includegraphics[width=15cm,height=6cm,trim=100 50 300 50,page=5]{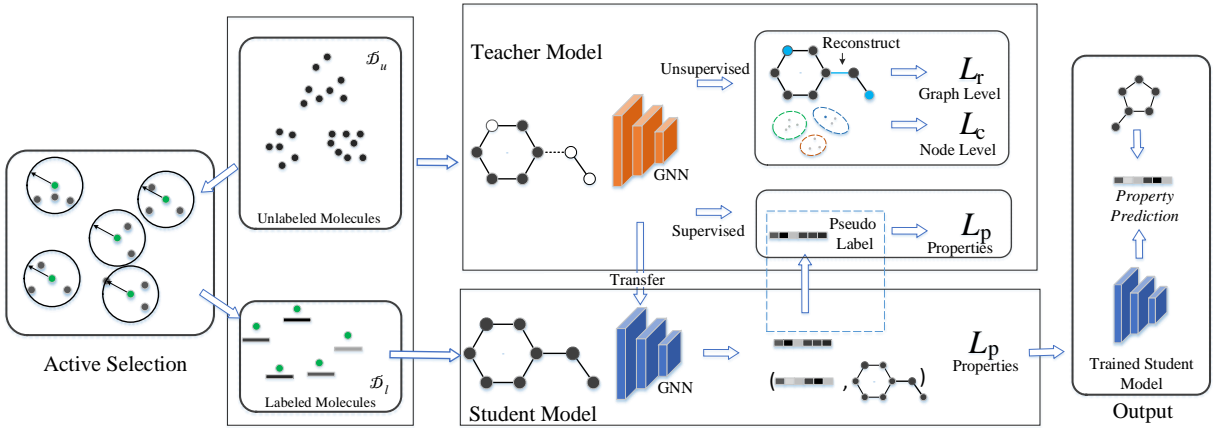}
	\vspace{-0.5cm}
	\caption{The overall framework of our method. We use a teacher model that jointly learns node embeddings at the local level and the distribution of the data at the global level with the property task. A student network uses the teacher's weight to fine-tune network parameters on property prediction task. Active learning and pseudo labeling are used to combine these steps effectively into a framework.}
	\label{fig2}
	\vspace{-0.5cm}
\end{figure*}

\section{ASGN: Active Semi-supervised Graph Neural Network}
In this section, we present a description of the framework of ASGN. Then we describe the components of ASGN comprehensively.
\vspace{-0.1cm}
\subsection{Framework}
In this paper, we propose a novel \textbf{A}ctive \textbf{S}emi-supervised \textbf{G}raph Neural \textbf{N}etwork (ASGN) for molecular property prediction by incorporating both labeled and unlabeled molecules in chemical space. The general framework is illustrated in Figure \ref{fig2}.

Generally, we use a teacher model and a student model that work iteratively. Each of them is a graph neural network. In the teacher network, we use a semi-supervised fashion to obtain a general representation of molecular graphs. We jointly train the embeddings for unsupervised representation learning and property prediction. Then in the student model, we handle the loss conflict by fine-tuning the parameters transferred from the teacher model for property prediction. After that, we use the student model to assign pseudo labels for the unlabeled dataset. As feedback for the teacher, the teacher model can learn the student's knowledge from these pseudo labels. Also, to improve the labeling efficiency, we propose using active learning to select the new representative unlabeled molecules for labeling. We then add them to the labeled set and finetune two models iteratively until accuracy budget is reached.
Specifically, the key idea is to use the embeddings output by the teacher model to find a subset that is most diversified in the whole unlabeled set. We then assign ground truth labels such as using DFT calculation to these molecules. After that, we add them into the labeled set and repeat the iteration to improve performance.

In the following,  we will first describe technical details of our teacher model and student model.
\vspace{-0.1cm}
\subsection{Semi-supervised Teacher Model }
In the teacher model, we use semi-supervised learning. We first introduce the network backbone. Then, we introduce the loss for representation learning. Specifically, a property loss on labeled molecule $\mathcal{D}_l$ and two unsupervised loss (from both the graph and the node level) on all molecules $\mathcal{D}_u \cup \mathcal{D}_l$ are designed to guide it. 

\subsubsection{Message Passing Graph Neural Network}
The task of the teacher model is to learn a general representation for molecular graphs from both labeled set and unlabeled set. We first introduce a message passing graph neural network (MPGNN) as the backbone that transforms a molecular graph into a representation vector based on message passing graph neural networks.
The graph neural network consists of $L$ message passing layers.
At $l$-th layer, it first embeds each node in a graph to a high dimensional space as their embeddings using $f(v_i)=z_i \in \mathbb{R}^d$. Then the node embeddings are updated by aggregating node embeddings of its neighbors $\mathcal{N}(v_i)$ along the weighted edges called message passing:

\begin{equation}
\boldsymbol{z}^{l + 1}_i = \sigma(\boldsymbol{W}^l \cdot \tmop{AGG} (\boldsymbol{z}_i^l, \{ \boldsymbol{e} (v_i, v_j) : v_j \in \mathcal{N} (v_i) \})),
\label{eq1}
\end{equation}
where $\sigma(\cdot)$ is the activation function, $\boldsymbol{W}^l$ is a learnable weight matrix, $AGG$ is the aggregation  function such as sum , mean, max \cite{ma2019graph}. Here we choose sum as the aggregation type which directly adds the messages from its neighbors as suggested in \cite{Xu2018}. $\tmmathbf{e}(v_i, v_j)$ is a vector called message function determined by the node embeddings and edge weights that pass from node $v_i$ to $v_j$. As the interactions decay with the growth of the distances between two atoms, we use a Gaussian radical basis \cite{schutt2017schnet} to embed the edge information that reflects the interaction strength between nodes:
\begin{equation}
\tmmathbf{e} (v_i, v_j)[k] = \boldsymbol{z}_{i}^{l} [k] \cdot \exp (- \gamma (\|
\tmmathbf{r}_i -\tmmathbf{r}_j \| - d_k)^2),
\label{eq2}
\end{equation} 
for $1 \leq k \leq N_f$ where $\{d_k: 1\leq k \leq N_f\}$ is a set of pre-defined filter centers. More intensive centers
means higher resolution and can capture minor difference of different bond
length. 

After $L$ layers of message passing and aggregation, we aggregate all node
embeddings to get the whole graph embedding:
\begin{equation}
\tmmathbf{z}_{\mathcal{G}} = \tmop{Pool} (\{ \tmmathbf{z}_{i}^{L} : v_i \in
\mathcal{V} \}).
\label{eq4}
\end{equation}
In this paper, we utilize a simple pooling method which directly averages or sums all
node embeddings. At last, multi-layer perceptron $f_\theta$ is used to get the  property
$f_{\theta} (\tmmathbf{z}_{\mathcal{G}})$.

Traditionally, MPGNN is trained in a supervised manner where all the labels are given and we usually use mean square loss (MSE) between predictions and labels $\boldsymbol{y}_i$ (i.e. the labeled properties in $\mathcal{D}_l$ ) to guide the optimization of the model parameters:
\begin{equation}
\mathcal{L}_p = \sum_{i = 1}^{N_l} \| \tmmathbf{y}_i - f_{\theta}
(\tmmathbf{z}_{\mathcal{G}_i}) \|^2 .
\label{eq5}
\end{equation}
However, in practice the training set with small number of labels easily results in an over-fitted model. Additionally, end-to-end training that only learns a high-level representation guided by the property/label is less effective for structural representation. To overcome these challenges, in this paper we propose a semi-supervised representation learning method by considering both local level and global level unsupervised information to enhance the expressive power of a model for both labeled and unlabeled molecular graphs.
\subsubsection{Node Level Representation Learning}

In node level representation learning, we learn to capture domain knowledge from geometry information of a molecular graph. The main idea is to use node embeddings to reconstruct the node types and topology (distances between nodes) from the representation. Specifically, we first randomly sample some nodes and edges from the graph as shown in Figure \ref{fig2}, then
pass these nodes' representation to a MLP and use them to reconstruct the node types $\boldsymbol{f}_{i}$ and 
distances between nodes $e_{i j}$. 
Mathematically, we  minimize the following cross-entropy loss:
\begin{equation}
\begin{aligned}
&\mathcal{L}_r = 
-\mathbb{E}_{v_i \sim \mathcal{V}}  \left[ \sum_{m =
	1}^{K_n} f_{i m} \log (g_{\theta_n} (\tmmathbf{z}_i)) \right] \\
&- 
\mathbb{E}_{e_{i j} \sim \mathcal{E}} \left[ \sum_{m = 1}^{K_e} e_{i j m}
\log (g_{\theta_e} (\tmmathbf{z}_i, \tmmathbf{z}_j)) \right], 
\end{aligned}
\label{eq6}
\end{equation}
where first term is the loss function for node types reconstruction, and the second term is the edge weights reconstruction. For both terms, we optimize the expectation of the samples. $K_n$ is the number of atom types, we transform the continuous edge weights into a discrete classification problem by dividing the continuous distance into several discrete bins and $K_e$ is the total number of bins. It means that 
$e_{i j m} = 1$ only if $d_m$ is \ the nearest to the weight of edge $e_{i j}$.
$g_{\theta_{n}}, g_{\theta_{e}}$ is a multi-layer perceptron. 

Practically, we randomly sample some nodes and edges to reconstruct their attributes and optimize the expectation of samples. 
We found such random sampling to be significantly more efficient without sacrificing much performance. 
We sample $\alpha |\mathcal{G}|$ ($0<\alpha<1$) edges from the graph along with the nodes to reconstruct their features. What's more, we notice that using a fully connected graph to represent a molecule contains redundant information because a molecule contains only $3n$ degrees of freedom since the coordinates of each atom can be decided by $3$ numbers as $(r^{(1)}, r^{(2)}, r^{(3)})$.
Therefore sampling edges with size $O(|\mathcal{G}|)$ is an efficient trade-off between performance and algorithm complexity. By optimizing the reconstruction loss (Eq. (\ref{eq6})), we can obtain the node embeddings that contains the topology and features of molecular graphs. 

\subsubsection{Graph Level Representation Learning}
Although node embeddings that can reconstruct the topology of molecules can effectively represent the structure of molecules, a recent study \cite{Hu} shows that combing graph level representation learning is beneficial for downstream tasks like property prediction. In order to learn a graph level representation, the key insight is to use the mutual relation between molecules within the chemical space, i.e. similar molecules roughly have similar properties. Inspired by this intuition, we propose a method based on learning to cluster to enhance graph level representation. 
First, we calculate the graph level embedding by the network. Then we use an implicit clustering based method to assign $N$ molecules each with a cluster id which contains $M$ clusters generated by the implicit clustering process. After that we optimize the model with a penalty loss function. The process is iteratively done until at least a local minima is reached. 

Next, we introduce the details of graph level representation learning. We denote $s$ as the cluster id in the rest of this section. First we pass the graph level embedding into a multi-layer perceptron and predict the probability distribution $p(s|\mathcal{G})$. We assume there exists a posterior distribution $p(s|\mathcal{G})$ of cluster id. We optimize the cross-entropy loss between $p$ and $q$ as following: 

\begin{equation}
H (y, x) = - \sum_{i = 1}^N \sum_{j = 1}^M p (s_j | \nobracket \mathcal{G}_i) \log q (s_j
| \nobracket \mathcal{G}_i).
\label{eq8}
\end{equation} 

However, we easily get a trivial solution if no constraint is applied on $p(s|\mathcal{G})$. The key is to confine these clustering ids to a pre-defined prior distribution $p(s)$ as $ \sum_{i=1}^{N} p(s_j|\mathcal{G}_i) = p(s_j)$ \cite{bojanowski2017unsupervised} \cite{asano2019self}. We choose a uniform distribution with fixed $M$ supports which means that the whole dataset is roughly divided into equally partitioned subsets. Practically, we use hard labeling technique to constraint $p(s|\mathcal{G}_i)$ to be a discrete label by applying the hardmax function. Then we explicitly write the optimization object as:

\begin{equation}
\min_{p, q}\mathcal{L}_c = \sum_{i = 1}^N \sum_{j = 1}^Mp(s_j|
\nobracket \mathcal{G}_i) \log q (s_j | \nobracket \mathcal{G}_i)
\label{eq9}
\end{equation}
\[
\tmop{s.t} : p (s_j | \mathcal{G}_i \nobracket) \in \{ 0, 1 \} ,  \sum_{i = 1}^N p (s_j | \mathcal{G}_i \nobracket) = p (s_j).
\]

We iteratively optimize predictive distribution $q(s|\mathcal{G})$ by performing gradient descent on the network parameters and the posterior distribution $p(s|\mathcal{G})$ by the following method which can be viewed as an implicit clustering approach. We first rewrite Eq. (\ref{eq9}) as: 

\begin{equation}
\min \mathcal{L}_c = \min_{Q \in U (p, q)} \langle P, Q \rangle, 
\label{eq10}
\end{equation} 
with $\langle \cdot, \cdot \rangle$ denotes the Frobenius dot-product
between two matrices, $P_{i j}=p(s_j|\mathcal{G}_i)$, $Q_{i j}=q(s_j|\mathcal{G}_i)$, and $U(p,q)$ denotes the joint distribution of $p$ and $q$. This is a typical optimal transport problem and we add
an entropy regularization and use Sinkhorn-Knopp  algorithm \cite{cuturi2013sinkhorn} for a better
convergence speed:
\begin{equation}
\min \mathcal{L}_c = \min_{Q \in U (p, q)} \langle P, Q \rangle -
\frac{1}{\lambda} \tmop{KL} (Q | | p q^T).
\label{eq11}
\end{equation}
In fact, this process can be viewed as a type of clustering \cite{cuturi2014fast} so we name this loss as clustering loss for self-supervision.

In a nutshell, to train a teacher model under a semi-supervised manner, we need to optimize the following loss jointly combining Eq. (\ref{eq5}), Eq. (\ref{eq6}) and Eq. (\ref{eq10}) as:
\begin{equation}
\mathcal{L}_t = \sum_{\mathcal{G} \in \mathcal{D}_l} \mathcal{L}_p + \sum_{\mathcal{G} \in \mathcal{D}_u \cup \mathcal{D}_l} \mathcal{L}_r + \sum_{\mathcal{G} \in \mathcal{D}_u \cup \mathcal{D}_l} \mathcal{L}_c.
\label{eq12}
\end{equation}

\subsection{Supervised Student Model}

Practically, directly optimizing Eq. (\ref{eq12}) of the teacher model yields unsatisfactory results for property prediction. 
The teacher model will be heavily loaded since it requires to learn several tasks simultaneously. Due to the conflict of these optimization targets, we observe that each target gets worse performance compared with optimizing them separately. Especially, it is also inefficient because if $|\mathcal{D}_l|<<\mathcal{D}_u$ then little attention will be paid to optimization of $\mathcal{L}_p$ in an epoch, however property prediction is what we care the most. 
As a result, the property prediction loss is much higher compared with a model that only needs to learn this task. To alleviate this problem, we propose introducing a student model. We use the teacher model to learn such representation by jointly optimizing the objects above. When the teacher's learning process ends, we transfer the teacher's weight to the student model, and use the student model to fine-tune only on the labeled dataset to learn the target properties the same as Eq. (\ref{eq5}) shown in Figure \ref{fig2}:
\begin{equation}
\mathcal{L}_s = \sum_{\mathcal{G}_i \in \mathcal{D}_l} ||\boldsymbol{y}_i - f_{\theta_s} (z_{\mathcal{G}_i})||^2.
\label{eq16}
\end{equation}
After fine-tuning, we use the student model to infer the whole unlabeled dataset and assign each unlabeled data a pseudo label indicating the student's prediction of its properties then the unlabeled dataset is $\mathcal{D}_u = \{(\mathcal{G}_i, f_{\theta_s}(\mathcal{G}_i)): 1\leq i \leq |\mathcal{D}_u|\}$ where $\theta_s$ is the parameters of student model. 
In the next iteration, the teacher model also needs to learn such pseudo labels as Eq. (\ref{eq12}) becomes:
\begin{equation}
\mathcal{L} = \sum_{\mathcal{G} \in \mathcal{D}_u \cup \mathcal{D}_l} \mathcal{L}_p + \sum_{\mathcal{G} \in \mathcal{D}_u \cup \mathcal{D}_l} \mathcal{L}_r + \sum_{\mathcal{G} \in \mathcal{D}_u \cup \mathcal{D}_l} \mathcal{L}_c.
\label{eq13}
\end{equation}

This can be viewed as the teacher learns the knowledge from the students as feedback inspired by the idea of knowledge distillation \cite{hinton2015distilling}. In summary, we handle the loss conflict by using two models whose targets are different. The teacher model learns a general representation while the student model aims to learn accurate prediction of molecular graph properties. The pre-training of the teacher provides a warm start for the student model.

\subsection{Active Learning for Data Selection}
We have incorporated the information in both labeled and unlabeled molecules. However, due to the limited number of labels available, the accuracy might still be unsatisfactory, we need to find new labeled data to improve its performance. Therefore, in each iteration we use the embeddings output by the teacher model to iteratively select a subset of molecules, and the properties (ground truth labels)  will be computed (i.e., by DFT). Then we add these molecules output by active learning into the labeled set for finetuning two models iteratively. Along this line, the key strategy of active learning is to find a small batch of most diversified molecules in the chemical space for labeling. A well-studied method to measure diversity is to sample from  $k$-DPP  as \cite{kulesza2011k} suggests. However, the subset selection is NP-hard therefore a greedy approximation is taken advantage of, which is the $k$-center method. Denoting the unlabeled dataset by $\mathcal{D}_u$, and the labeled dataset by $\mathcal{D}_l$, we use a myopic method that in each iteration we choose a subset of data that maximize the distance between labeled set and unlabeled set.
Concretely, for every $0 < i < b$ within the $k$-th batch, we choose the data point that
satisfies the following condition:
\begin{equation}
\tmop{argmax}_{j \in [n] \backslash\mathcal{D}_u^k} \min_{i \in
	\mathcal{D}_l^k} d (\mathcal{G}_i, \mathcal{G}_j),
\label{eq14}
\end{equation}
where $d(\mathcal{G}_i,\mathcal{G}_j) = \| \boldsymbol{z}_{\mathcal{G}_i} - \boldsymbol{z}_{\mathcal{G}_j}\|$ is the distance between two molecules. We use $L-2$ norm on the representations by the teacher model. Since the teacher model learns a general representation we naturally believe that the distance between the representations of two molecules indicates the difference of them. Moreover the features are automatically extracted, we do not need to rely on handcraft distances like graph edit distance which might not suit our problem. Additionally, since the teacher model is trained in a semi-supervised manner, the teacher model only needs to be fine-tuned when new labeled data is added, thus accelerating the training process.

\vspace{-0.2cm}
\subsection{Method Summary and Discussion}
In this subsection, we briefly summarize the framework in Algorithm \ref{al1}. Given a unlabeled set and a labeled set. In each iteration, we use $k$-center active learning strategy to get a new batch of data for labeling and add them to the labeled set (Line 4), next we transfer the teacher's weight to the student network (Line 5) and fine-tune the student network (Line 6),  then we use the student model to assign a pseudo label of the property for the rest of the unlabeled dataset (Line 7). After that, we continue to fine-tune the teacher model jointly with three tasks (Line 8). At last, the trained student model will be applied to predict the properties of the molecules.

\begin{algorithm}
	\caption{ASGN framework}
	\label{al1}
	
	\textbf{Input:}Unlabeled,labeled,test dataset $\mathcal{D}_u$,$\mathcal{D}_l$,$\mathcal{D}_{test}$, error $\epsilon(\cdot, \mathcal{D})$, batch size $b$, stopping error,label budget $\epsilon$,$B$ 
	
	\textbf{Output:}student model $\theta_s$ 
	\begin{algorithmic}[1]
		
		\STATE Initialize teacher and student $\theta_t$,$\theta_s$, labeled dataset $\mathcal{D}_l$ 
		
		\WHILE{$\epsilon(\theta_s,\mathcal{D}_{test})>\epsilon$ or $|\mathcal{D}_l| \leq B$}  
		
		\STATE Pre-train/finetune the teacher model by minimizing $\mathcal{L}=\mathcal{L}_r+\mathcal{L}_c+\mathcal{L}_p$ to get graph embeddings $\{z_{\mathcal{G}}:\mathcal{G} \in \mathcal{D}_u\}$.
		
		\STATE Use $k$-center active learning with $\boldsymbol{z}_\mathcal{G}$ for querying new labeled data $s$, $\mathcal{D}_l \gets \mathcal{D}_l \bigcup s$, $|s|=b$. 
		
		\STATE Transfer the weights of teacher to student $\theta_s \gets \theta_t$. 
		
		\STATE Finetune the student network by minimizing $\mathcal{L}_p = \epsilon(\theta_s,\mathcal{D}_l)$.
		
		\STATE Assign pseudo label for the unalabeled dataset using student model, $\boldsymbol{y}_i \gets f_{\theta_s}(\mathcal{G}_i)$,$i \leq |\mathcal{D}_u \setminus \mathcal{D}_l|$.

		\ENDWHILE
		
		\textbf{Return:} student model $\theta_s$
		
	\end{algorithmic}
\end{algorithm}

To summarize,  we propose a novel approach to predict the properties of molecules using graph neural networks.
First, we use a multi-level representation learning method to obtain general embeddings for molecular graphs. The node embeddings store  essential components of molecular graphs and they are composable to form meaningful graph level embeddings with respect to the whole data distribution. Subsequently, a teacher-student framework is used to effectively combine semi-supervised learning and active learning to deal with label insufficiency. Compared with vanilla semi-supervised learning methods \cite{Sener2017}, the separation of the two models can alleviate loss conflict. Compared with naive active learning methods that re-trains the model from scratch when every new batch data points are selected, the weight transferred from the teacher provides a warm start for the student and avoids overfitting of the small labeled dataset and accelerates training. Besides, the two models communicate via weight transfer and feedback from assigning pseudo labels so that they can be mutually promoted.

\section{Experiments}

\begin{table*}[]
	\centering
	\caption{Results on QM9 dataset for effectiveness experiment.}
	\vspace{-0.4cm}
	\begin{tabular}{l|llllllllllll}
		\hline
		Properties & $U_0$ & $U$ & $G$ & $H$ & $C_v$ & HOMO & LUMO & gap & ZPVE & $R^2$ & $\mu$ & $\alpha$ \\ \hline
		Unit & eV & eV & eV & eV & Cal/MolK & eV & eV & eV & eV & Bohr$^2$ & Debye & Bohr$^3$ \\ \hline
		Supervised & 0.3204 & 0.2934 & 0.2948 & 0.2722 & 0.2368 & 0.1632 & 0.1686 & 0.2475 & 0.0007 & 10.05 & 0.3201 & 0.5792 \\
		Mean-Teachers & 0.3717 & 0.2730 & 0.2535 & 0.2150 & 0.2036 & 0.1605 & 0.1686 & 0.2394 & 0.00054 & 5.22 & 0.3488 & 0.5792 \\
		InfoGraph & 0.1410 & 0.1702 & 0.1592 & 0.1552 & 0.1965 & 0.1605 & 0.1659 & 0.2421 & 0.00036 & 4.92 & 0.3168 & 0.5444 \\
		ASGN (Ours) & \textbf{0.0562} & \textbf{0.0594} & \textbf{0.0560} & \textbf{0.0583} & \textbf{0.0984} & \textbf{0.1190} & \textbf{0.1061} & \textbf{0.2012} & \textbf{0.00017} & \textbf{1.38} & \textbf{0.1947} & \textbf{0.2818} \\ \hline
	\end{tabular}
	
	\label{tab:qm91}
	\vspace{-0.2cm}
\end{table*}

\begin{table}[]
	\footnotesize
	\caption{Results on OPV dataset for effectiveness experiment.}
	\vspace{-0.3cm}
	\begin{tabular}{l|ll}
		\hline
		Property & HOMO & LUMO \\ \hline
		Unit & \multicolumn{2}{c}{Hatree} \\ \hline
		Supervised & 0.080 & 0.078 \\
		Mean-Teacher & 0.078 & 0.075 \\
		InfoGraph & 0.077 & 0.076 \\
		ASGN (Ours) & \textbf{0.059} & \textbf{0.057} \\ \hline
	\end{tabular}
	
	\label{Opvexp1}
	
\end{table}

In this section, we conduct extensive experiments to show the effectiveness of ASGN on two popular molecular datasets. The code is publicly available \footnote{https://github.com/HaoZhongkai/AS\_Molecule}.
\vspace{-5pt}
\subsection{Datasets}

\begin{itemize}
	\item \textbf{QM9:\footnote{http://quantum-machine.org/datasets/}} The QM9 dataset \cite{ramakrishnan2014quantum} is a well-known benchmark datasets that contains the equilibrium coordinates of 130,000 molecules along with their quantum mechanical properties. We use 10,000 molecules for testing and 10,000 for validation. 
	Coordinates and properties for all molecules are calculated using DFT methods. Molecules in QM9 contain no more than 9 heavy atoms (atom heavier than hydrogen).
	\item \textbf{OPV:\footnote{https://cscdata.nrel.gov/\#/datasets/ad5d2c9a-af0a-4d72-b943-1e433d5750d6}}
	OPV \cite{st2019message}is a dataset with roughly 100,000 medium size molecules, each contains 20 to 30 heavy atoms. Again the properties and equilibrium coordinates of these molecules are obtained through DFT. We use 5,000 for testing and 5,000 for validation.
\end{itemize}

\vspace{-4pt}
\subsection{Experiments Setup}
We evaluate our method under two experimental settings. We first describe the implementation details and parameters of ASGN. We run all experiments are on one Tesla V100 GPU and 16 Intel CPUs.

\textit{Graph Neural Network Hyperparameters}.
For the network backbone, we use 4 message passing layers and embedding dimension of 96 in Eq. (\ref{eq1}). We use Adam optimizer with a learning rate 1e-3. We use filters from 0 to 3nm with an interval of 0.01nm in Eq. (\ref{eq2}).

\textit{Semi-Supervised Learning Hyperparameters}.
The teacher model has an additional linear classifier after the graph neural network. We divide the distance of the edge into $30$ bins in Eq. (\ref{eq6}) for reconstruction. We use $M=100$ in Eq. (\ref{eq8}). The regularization constant $\lambda$ is set to be 25 in Eq. (\ref{eq11}). We train (fine-tune) the teacher model for 20 epochs in each iteration. We train the student network until the loss does not decrease for about 20 epochs.

\textit{Active Learning Hyperparameters}.
In each iteration, we select 1,000 new unlabeled molecules in Eq. (\ref{eq14}) to be labeled and add them into the training dataset.

\begin{figure}
	
	\includegraphics[width=9cm,height=4cm,trim=40 90 90 170,clip,page=2]{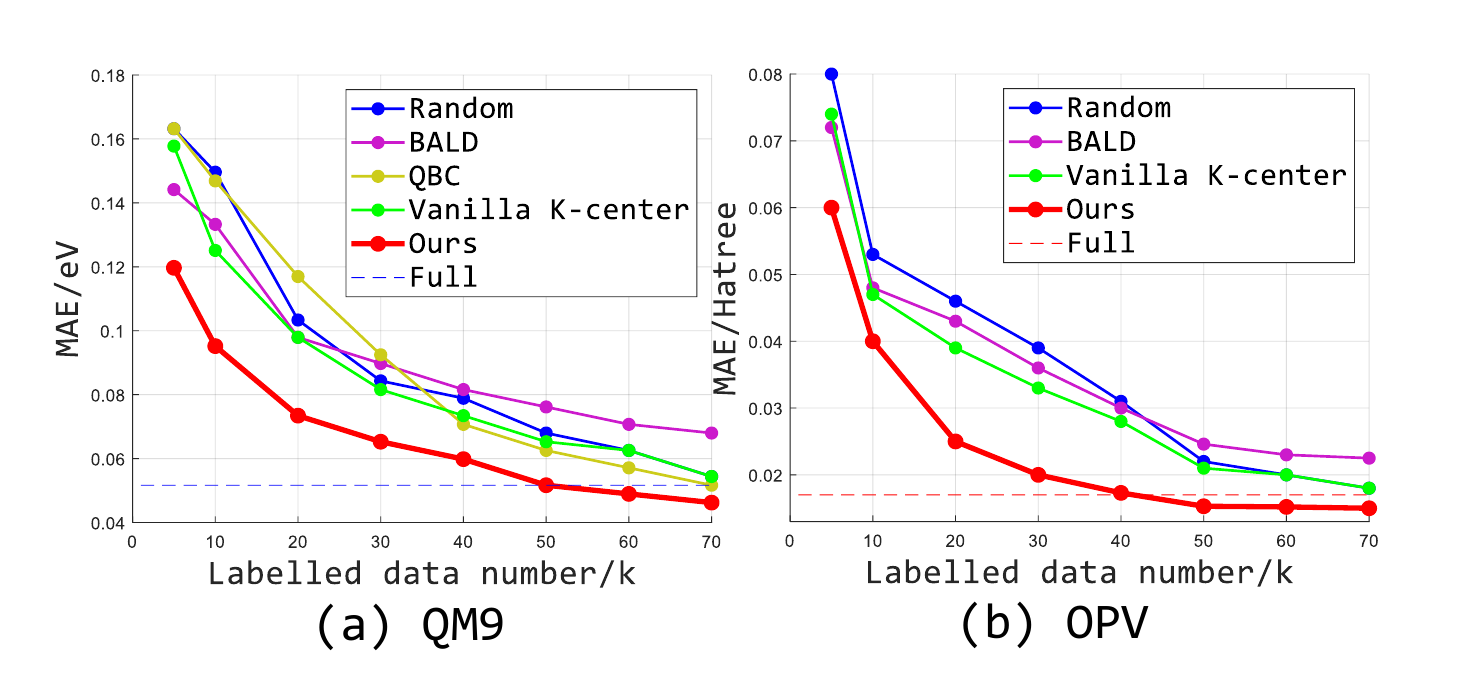}
	\vspace{-0.6cm}
	\caption{The results of efficiency experiments of property HOMO on QM9 and OPV datasets. }
	\label{free1}
	\vspace{-0.4cm}
\end{figure}

\subsection{Effectiveness Experiment}
To demonstrate that our method could achieve lower error with limited labeled data, we first conduct an effectiveness experiment. Under this experimental setting we have a fixed label budget which is the maximum number of labels. Given a fixed label budget, we compare the final Mean-Absolute-Error(MAE) \cite{schutt2017schnet} on the test dataset after training. We use a label budget of 5000 for both QM9 and OPV about 5\%. Other than these 5,000 labeled data, other labels are not available.
We compare our methods with baselines listed below.

\subsubsection{Baselines}
For accuracy experiments, we mainly compare our method with several semi-supervised learning baselines. To ensure fairness, all baselines are conducted on the same network backbone (i.e MPGNN). The compared baselines are selected from two perspectives, one is traditional semi-supervised learning, the other is semi-supervised learning baselines for graph data.

\begin{itemize}
	\item \textbf{Supervised :} We train the network backbone using fully supervised manner only on the small labeled dataset.
	
	\item \textbf{Mean-Teachers \cite{tarvainen2017mean}:} This is a method for semi-supervised learning by using a consistency regularization and uses moving average for the models' weights as the teacher. 
	
	\item \textbf{InfoGraph \cite{sun2019infograph}:} This is the state-of-the-art method for  semi-supervised learning or unsupervised learning on graphs. It maximizes the mutual information between the graph level representations and the substructures of the graphs.

\end{itemize}

\vspace{-0.4cm}
\subsubsection{Results}
The results are listed in Table \ref{tab:qm91} on QM9 dataset and Table \ref{Opvexp1} on OPV dataset.

First, We found that our method is significantly better than baseline methods on all properties. We achieved a reduction of more than 50$\%$ on several properties such as $U$, $U_0$, $\alpha$ and $C_v$ compared with the state-of-the-art method. This shows our semi-supervised learning method is effective and incorporating unlabeled data can help the prediction of molecular properties.

Second, the semi-supervised reconstruction captures domain knowledge for molecules and achieves better results than supervised model (i.e MPGNN) and Mean-Teachers. The global representation learning at graph level is beneficial for molecular property prediction and its performance is better than Infograph.

\subsection{Efficiency Experiment}
To demonstrate ASGN is label efficient, we conduct an efficiency experiment.
In this experiment, we start with 5,000 labeled molecules and the rest in the unlabeled set. Then, in each iteration, after the model selects a molecule from $\mathcal{D}_u$, we add it to $\mathcal{D}_l$.
During this process, we measure the Label Rate-Mean Absolute Error(MAE) curve to show how many labels are saved for a fixed error. For a fixed error, the less labeled data is used, the better the model is.

\subsubsection{Baselines}
The baselines are selected from active learning methods. We apply these methods on the backbone of ASGN (i.e MPGNN). We simply omit some methods that cannot be applied to our settings. We use a batch number of 2500 new labeled molecules in every iteration in Eq. (\ref{eq14}) for ASGN. The computational cost of QBC method on OPV dataset is unaffordable so we simply omit it.
\begin{itemize}
	\item \textbf{Random:} Choosing data points randomly from the unlabeled dataset in each iteration. The model is re-initialized when a new batch of labeled data is selected. This method equals the passive learning.
	
	\item \textbf{Query By Committee (QBC) \cite{seung1992query}:} We jointly train a group of models named committee initialized in the same method but different parameters. Each iteration we choose a batch of data points with the biggest disagreement of the committee members. We use 8 models as a committee, training 8 models at the same time is time consuming.
	
	\item \textbf{Deep Bayes Active Learning (BALD) \cite{gal2017deep}:} This is a method based on uncertainty. We approximate the uncertainty by performing Monte Carlo dropout \cite{srivastava2014dropout} on layers of the network.
	
	\item \textbf{Vanilla $k$-center \cite{Sener2017}:} The representation learned by the semi-supervised learning methods actually benefits the selection of new data points. We also compare our method with the vanilla plain $k$-center active learning strategy.

\end{itemize}

\subsubsection{Results}
We plot the results on HOMO (highest occupied molecular orbital) on both QM9 dataset and OPV dataset in Figure \ref{free1}. "Full" denotes the MAE for a supervised MPGNN using all labeled data. We have the following conclusions. 

First, we show that for all datasets and properties, when the label number is fixed, the MAE is much lower than baselines which proves the effectiveness of our model. This shows that the active learning strategy is beneficial for model training. Additionally, the performance is better than a fully supervised model on all labeled data, proving the effectiveness of combining semi-supervised loss as regularization.

Second, when we set a fixed error target, we found that our model is  about $2\sim3$ times label efficient than baselines. This means that if we only need a predictor with given accuracy, we could use only $1/3\sim1/2$ labels compared with other methods. Specifically, we use 50\% labeled data to reach full accuracy on QM9 and 40\% for OPV.

Third, we found that some baseline methods that work well in deep learning for image classification like BALD and $k$-center do not perfrom well on molecular data. Additionally, since BALD requires dropout, the performance is better when few labels are available but worse when we use all the labels.

\subsection{Ablation Experiments}
In this section, we conduct more experiments on ASGN including the ablation study to demonstrate how every part of our model affects the performance and a visualization experiment to support the interpretability of our model.

\subsubsection{Necessity of Teacher-Student Framework}
First, to show the effectiveness of the teacher-student framework in our model, we conduct an ablation study of ASGN without the teacher model or the student model.
We denote ASGN with only the teacher model as ASGN-T which means that we jointly learn all tasks without handling the loss conflict. We list the results of HOMO on QM9 and OPV datasets in Table \ref{tab:ab}. We see that with the student network, the model achieves better performance on property prediction task.

\begin{figure}
	\vspace{-0.6cm}
	\centering
	\includegraphics[width=7cm, height=5cm,trim=50 50 50 50,page=6]{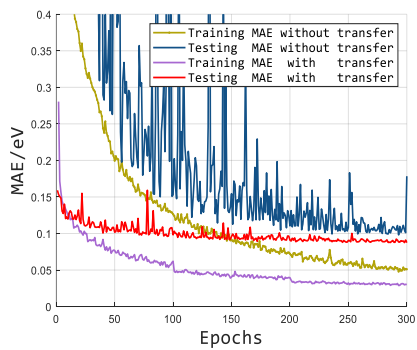}
	\caption{Ablation resuts on the necessity of weight transfer.}
	\label{exp2}
	\vspace{-0.7cm}
\end{figure}
We also study the case without the teacher model as ASGN-S which means no semi-supervised learning is used. Notice that ASGN-S is identical to a vanilla $k$-center active learning method\cite{Sener2017}. Results show that it is necessary using the teacher-student framework.

\subsubsection{Necessity of Weight Transfer}
The essential step in connecting the student model and teacher model in our method is to transfer the weight of the teacher model to the student model in order to accelerate the training process. Here we use an ablation experiment to demonstrate the effect of the weight transfer. In Figure \ref{exp2}, we plot the MAE of ASGN with weight transfer and without weight transfer on the test dataset of QM9 on LUMO (lowest unoccupied molecular orbital) property when 10,000 labeled data are available. Results show that both training and testing MAE converge faster and are more stable with weight transfer. The final performance is also better using weight transfer.

\subsection{Visualization Experiments}

\begin{figure}
	\vspace{-0.3cm}
	\centering
	\includegraphics[width=8cm, height=6cm,page=8,trim=50 120 30 10]{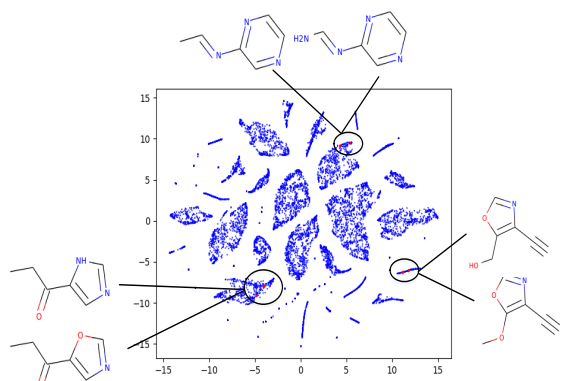}
	\vspace{-0.6cm}
	\caption{Visualization results on QM9 dataset of molecular graph embeddings using t-SNE method.}
	\label{vis}
	\vspace{-0.4cm}
\end{figure}

\begin{table}[]
	\footnotesize
	\caption{Results of Ablation experiments on the necessity of teacher-student framework.}
	\vspace{-0.2cm}
	\begin{tabular}{l|llllll}
		\hline
		Name/Dataset & \multicolumn{3}{c}{Homo(QM9)} & \multicolumn{3}{c}{Homo(OPV)} \\ \hline
		Unit & \multicolumn{3}{c}{eV} & \multicolumn{3}{c}{Hatree} \\ \hline
		Number of data & 5k & 10k & 50k & 5k & 10k & 50k \\
		ASGN-T & 0.1668 & 0.1523 & 0.0682 & 0.080 & 0.053 & 0.020 \\
		ASGN-S & 0.1632 & 0.1252 & 0.0653 & 0.076 & 0.049 & 0.019 \\
		ASGN & \textbf{0.1190} & \textbf{0.0951} & \textbf{0.0517} & \textbf{0.060} & \textbf{0.039} & \textbf{0.015} \\ \hline
	\end{tabular}
	
	\label{tab:ab}
	\vspace{-0.5cm}
\end{table}
Our representation learning has considered the mutual relation between molecules within the chemical space and we use the information mutually for predicting the clustering to enhance the representation. To demonstrate that the distribution of molecules exhibits a clustered structure, we use t-SNE method  to visualize the graph level representation of molecules using ASGN, shown in Figure \ref{vis}. We see after using t-SNE the embedding of molecules can be clustered, and there is obvious distance between the clusters which verifies that we have got discriminative graph level embeddings. Additionally, similar molecules are clustered into the same cluster that means the embeddings can capture structural information.

\vspace{-0.2cm}
\section{Conclusions}

In this paper, we proposed a novel framework to improve the performance for molecular property prediction with limited labels by incorporating unlabeled molecules. We designed a teacher-student framework consisting of two graph neural networks that work iteratively. Then we introduced the details of our semi-supervised representation learning method for molecular graphs that consider both graph level and node level information.  Weight transfer and pseudo labeling are used to optimize two models to balance the loss functions. Furthermore, we used diversity based active learning to select new molecules for labelling. ASGN achieves much better performance compared with baselines when labels are limited. Additionally, we showed the necessity for components in ASGN using ablation experiments. In future work, we will attempt to extend our model to more general molecular property prediction.

\textbf{ACKNOWLEDGMENTS}. This research was supported by grants from the National Natural Science Foundation of China (Grants No. 61922073, U1605251). Qi Liu gratefully acknowledges the support of the Youth Innovation Promotion Association of CAS (No. 2014299).

\footnotesize
\bibliography{ref.bib}
\bibliographystyle{ACM-Reference-Format}

\appendix

\end{document}